\documentclass[10pt,twocolumn,letterpaper]{article}

\usepackage{cvpr}              
\usepackage[accsupp]{axessibility}

\usepackage{graphicx}
\usepackage{amsmath}
\usepackage{amssymb}
\usepackage{booktabs}
\usepackage{bm}

\usepackage[pagebackref,breaklinks,colorlinks]{hyperref}
\usepackage[ruled]{algorithm2e} 
\usepackage{zref-xr}
\zexternaldocument[I-]{tex/9-Appendix}

\usepackage[capitalize]{cleveref}
\crefname{section}{Sec.}{Secs.}
\Crefname{section}{Section}{Sections}
\Crefname{table}{Table}{Tables}
\crefname{table}{Tab.}{Tabs.}

\def\cvprPaperID{16102} 
\def\confName{CVPR}
\def\confYear{2024}

\begin{document}

\title{
DiffusionPoser: Real-time Human Motion Reconstruction From Arbitrary Sparse Sensors Using Autoregressive Diffusion
}

\author{Tom Van Wouwe  \hspace{0.5cm} Seunghwan Lee  \hspace{0.5cm} Antoine Falisse \hspace{0.5cm}  Scott Delp \hspace{0.5cm}  C. Karen Liu\\
Stanford University\\
{\tt\small [tvwouwe,lsw9021,afalisse,delp,ckliu38]@stanford.edu}
}
\maketitle

\begin{abstract}
Motion capture from a limited number of body-worn sensors, such as inertial measurement units (IMUs) and pressure insoles, has important applications in health, human performance, and entertainment. Recent work has focused on accurately reconstructing whole-body motion from a specific sensor configuration using six IMUs. While a common goal across applications is to use the minimal number of sensors to achieve required accuracy, the optimal arrangement of the sensors might differ from application to application.
We propose a single diffusion model, DiffusionPoser, which reconstructs human motion in real-time from an arbitrary combination of sensors, including IMUs placed at specified locations, and, pressure insoles.
Unlike existing methods, our model grants users the flexibility to determine the number and arrangement of sensors tailored to the specific activity of interest, without the need for retraining. A novel autoregressive inferencing scheme ensures real-time motion reconstruction that closely aligns with measured sensor signals. The generative nature of DiffusionPoser ensures realistic behavior, even for degrees-of-freedom not directly measured. Qualitative results can be found on our \href{https://diffusionposer.github.io/}{website: https://diffusionposer.github.io/}.


\end{abstract}


\section{Introduction}
Portable and minimally intrusive tools for estimating human motion have important applications in health and human performance as they allow continuous monitoring of the motions of the body and the forces in the musculoskeletal system \cite{Uhlrich22,Lee22}. Mobile sensing technologies also have applications in virtual and augmented reality, and video games \cite{Schepers18,Schreiner21}. Truly wearable sensors such as inertial measurement units (IMUs) (e.g. \cite{al2022opensense}), pressure insoles (e.g. \cite{chatzaki2021smart}) or electromagnetic sensors (e.g. \cite{kaufmann2021pose}), have an advantage compared to video-based motion capture, as they are egocentric, do not suffer from occlusion or poor lighting, and offer an infinite measurement volume. While egocentric video represents an alternative or supplementary modality, it introduces potential privacy concerns.  With IMU sensors being the cheapest alternative of these mobile sensing technologies, much research has focused on motion reconstruction from these. 
\begin{figure}
  \includegraphics[width=1.0\linewidth]{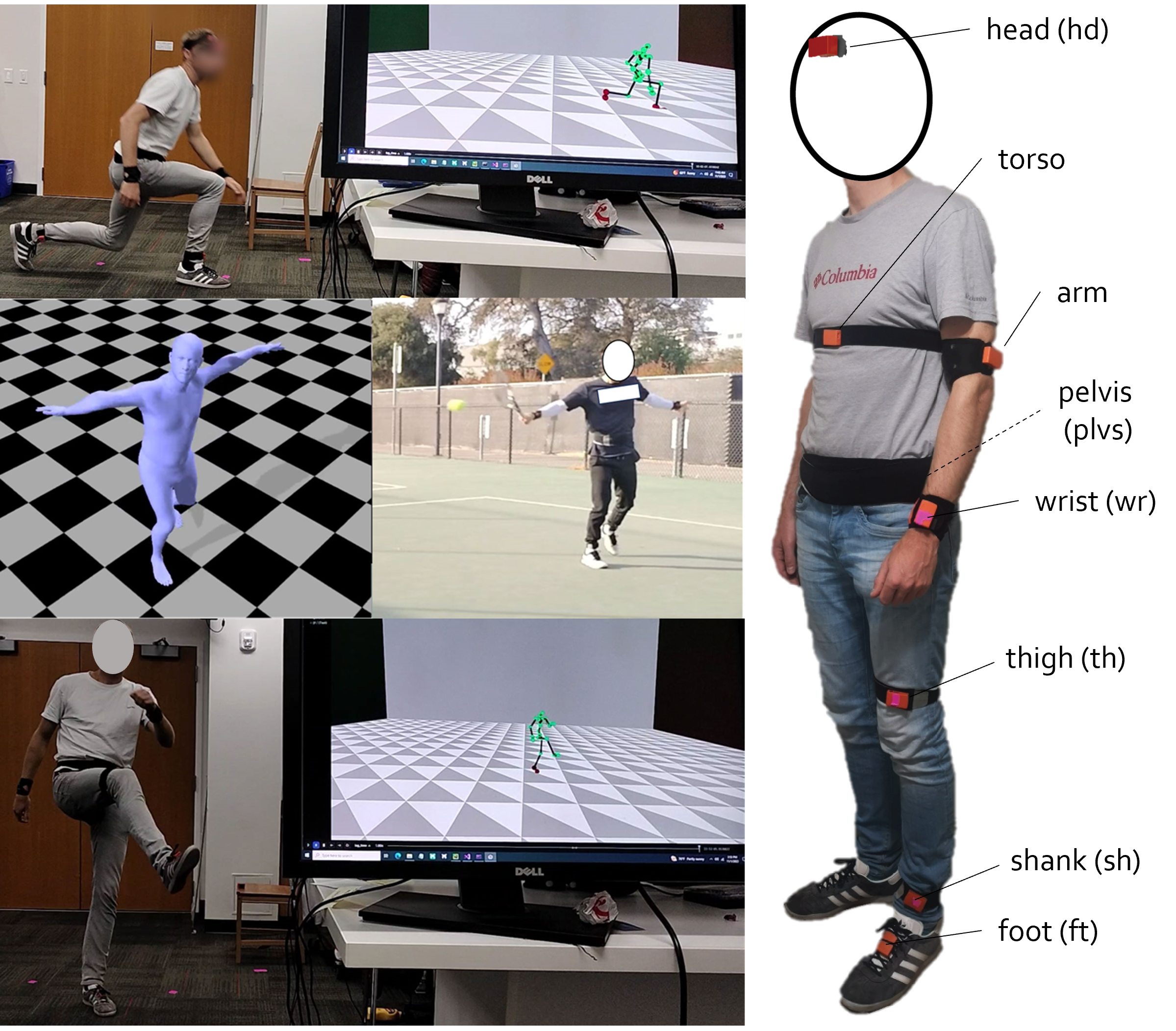}
  \caption{\label{fig:teaser} (Left) \textbf{Examples of live reconstruction using DiffusionPoser}. (Right) \textbf{Subject instrumented with IMUs}. We assume IMUs may be attached at 13 specific locations: pelvis, thighs, shanks, feet, arms, wrists, torso, head.}
\end{figure}
Configurations with few sensors, i.e. sparse configurations, are desirable for user convenience and adherence. However, a sparse configuration as opposed to a dense configuration, does not allow direct measurement of the full body motion. Besides sparsity, sensor noise is an important challenge. To tackle the problems of sparsity and noise in the case of IMUs, both optimization \cite{VonMarcard17} and data-driven \cite{Huang18, Slyper08} methods have been used.

Prior data-driven methods that reconstruct motion from sparse IMU sensor configurations focused on a specific number of sensors in a specific configuration \cite{Slyper08,VonMarcard17,Huang18,Yi21,Yi22,Jiang22}. In practice, applications would benefit from a real-time system that is more flexible and where the user can optimize the number of sensors and configuration to the activity of interest.

Here we present a single diffusion generative model, DiffusionPoser, that reconstructs human motion in real-time from ad-hoc sensor configurations, i.e. any number (between one and thirteen) and combination of IMUs and/or pressure insoles can be used. 

The use of a diffusion generative model is key to allow any combination of sensors. It results in realistic motion reconstruction in correspondence with measured signals, even for very sparse configurations. Our generative model is also robust to cases where sensor signals are corrupted or lost.

Two developments, that both utilize the inpainting denoising technique \cite{Lugmayr2022RePaintIU}, differentiate our model from previous motion diffusion models such as \cite{Tseng22, Tevet23, Li23, Du23} and enable our real-time sensor-based motion reconstruction application: 
\begin{enumerate}
    \item We use an \textbf{autoregressive inference scheme} (Section \ref{section:autoregressive_inference}) where at every time step the motion history is used to generate a full motion sequence, including the current frame of interest.
    \item We use \textbf{a tailored motion representation and inpainting denoising},  rather than a conditional model, to control motion generation. The motion representation consists of contact labels, root translation, global segment orientations and global linear accelerations of segment-specific sites.
\end{enumerate}

We demonstrate the capability of DiffusionPoser to optimize sensors configurations for applications that might require different levels of accuracy for specific body regions. We show that, in the case of using three sensors, the best configuration instruments, the pelvis and wrists when optimizing whole-body reconstruction, the pelvis and thighs when optimizing reconstruction of leg motion, and the upper arms and pelvis when optimizing reconstruction of the back, neck and shoulder joints. All these configurations can be used without retraining. 
We show that such flexibility does not come at the cost of accuracy as DiffusionPoser achieves accuracy that is on par with state-of-the-art regressive models that focus on a single specific six sensor IMU setup \cite{Huang18,Yi21,Yi22,Jiang22}.  

Our method is agnostic to the underlying skeleton models and we present two versions of DiffusionPoser for different skeletons. First, we use the SMPL body model \cite{Loper15}. Second, because we aim to enable biomedical research applications, we implemented DiffusionPoser for a different and more physiologically realistic skeleton: the OpenSim musculoskeletal model \cite{Seth18}. 
While DiffIP for the SMPL model was trained on the available AMASS dataset we created a new dataset specific to OpenSim models by combining three existing motion capture datasets \cite{Joo15,Szczkesna21,Trumble17}.

DiffusionPoser has interesting features for health and biomedical research applications. It reconstructs motion from wearable IMU sensors that have extended battery lives and are portable. It allows users to select a sensor configuration while trading off comfort and specific accuracy that suits their goal best, on-the-fly, without retraining. Finally, DiffusionPoser runs at 20Hz making it useful for health and performance interventions such as guidance of rehabilitation exercises or training using biofeedback \cite{Uhlrich22a}.

\section{Related Work}

\paragraph{Motion Capture using sparse IMUs.}
An IMU measures 3D linear acceleration, 3D angular velocity, and the direction of the magnetic North. These measurements are noisy and expressed in the local IMU frame. Commercially available IMU systems come with proprietary sensor fusion algorithms providing orientation and acceleration for use in downstream applications (e.g.,\cite{Roetenberg09}). Linear accelerations remain noisy and are therefore sometimes smoothed before being used as reconstruction input (e.g. \cite{Jiang22}).

Early work performing human motion reconstruction from sparse IMUs performed database search based on four \cite{Tautges11} or five accelerometer signals \cite{Slyper08}. In recent years, several works focused on reconstructing whole-body motion from six IMU sensors placed on the wrists (2), on the shanks (2), on the pelvis, and on the head. 
Sparse Inertial Poser \cite{VonMarcard17} relies on this configuration and does offline motion reconstruction by optimizing a sequence of poses to match measured signals. 

Deep Inertial Poser \cite{Huang18} introduced the use of a human motion prior: a bi-directional RNN architecture is trained on a motion dataset. IMU signals for training were synthesized from the AMASS motion dataset \cite{Mahmood19}. Other data-driven algorithms have built on this work to improve accuracy and address root translation estimation. Transpose \cite{Yi21} combines several RNNs to predict the full pose, including the root motion. An extension of this work, Physics Inertial Poser (PIP) \cite{Yi22}, adds a physics layer to refine the joint orientations and root motion predictions. Transformer Inertial Poser (TIP) \cite{Jiang22}, addresses root motion estimation by predicting stationary points on feet, hands and pelvis and applying a correction accordingly.

A downside of the prior work, that we address with DiffusionPoser, is that only one specific IMU configuration is allowed. One other prior work performs online reconstruction from a flexible and sparse IMU configuration: IMUPoser \cite{Mollyn23} uses an LSTM to predict motion from different IMU configurations including up to three sensors embedded in a phone, watch, and earbuds. Compared to our work, IMUPoser has a limited number of possible locations for the sensors and does not reconstruct root motion.

Finally, several systems complement inertial measurements with another sensory modality to improve reconstruction (e.g.: optical markers \cite{Andrews2016}, lidar \cite{Ren23}, third-person video \cite{Malleson17}, first-person video\cite{Yi23}, and depth camera \cite{Zheng18}). 

\paragraph{Diffusion for human motion reconstruction.}
Diffusion models \cite{Sohl15} are a class of deep generative
models. During training, a neural net learns the inverse mapping of samples from the target distribution that are gradually noised (i.e. denoising). At inference, the neural net is used to perform a denoising process starting from pure noise, thus generating a clean sample. A recent overview of diffusion models for human motion generation can be found in Section 7 of \cite{Po:2023:star_diffusion_models}.

A diffusion generative model can be controlled in different manners. One approach that has been employed in motion generative systems is to utilize a conditional generative model. In this model, a specific condition, such as text (e.g., \cite{Tevet23}) or music (e.g., \cite{Tseng22}), is provided during both the training and inference phases. Text-to-motion generative systems that rely on diffusion \cite{Tevet23,Kim22,Zhang22} have been shown to improve expressiveness and robustness compared to prior generative models \cite{Ahuja2019, Petrovich22} based on variational autoencoders \cite{Kingma2013}. 

Similar to our work, human motion diffusion models have been been used to reconstruct motion from different sensory modalities such as monocular camera \cite{Holmquist_2023_ICCV, gong2023diffpose}, egocentric camera \cite{Li23}, or position and orientation sensors \cite{Du23, Winkler22,Jiang22b}. There are two important distinctions between our work and the systems mentioned here. First, the mentioned systems have access to a good estimate of position as input, whereas we start from noisy linear acceleration signals. Second, the prior works are conditional models for which the sensor measurement is the condition of the generative process. We use the sensor measurement for guidance instead of using it as a condition. This allows to maintain flexibility with respect to what sensor signals we have access to.

Indeed, when using conditional models, we cannot use a different condition during inference than during training. Guidance on the other hand typically only interferes during inference. There are several ways to do guidance \cite{Po:2023:star_diffusion_models} and we take inspiration from \cite{Tseng22,Tevet23} that use inpainting denoising as a flexible approach to control (part of) the features of a generated motion. Different from how inpainting denoising is used in \cite{Tseng22} to generate long sequence motions by having small overlapping temporal patches, we generate a continuous motion sequence in a fully autoregressive manner. Each newly generate window of motion has an overlapping temporal patch with the previous window that covers all but the last frame.

\section{DiffusionPoser}
DiffusionPoser reconstructs whole-body human motion in real-time based on measurements from sparse IMUs in arbitrary configurations and/or pressure insoles.

\subsection{Skeleton model and IMU instrumentation}
We implemented DiffusionPoser for two body models: SMPL \cite{Loper15} and OpenSim \cite{Lai2017}. The system works identical for both models. Training is slightly different because of different definitions of the underlying kinematic tree. We provide further detail on the OpenSim model in Appendix A and focus on SMPL here.

The SMPL skeleton model is a 75 dof kinematic tree consisting of a root segment (6 dof) and 23 additional segments connected by ball-and-socket joints (3 dof). 
We assume that IMUs can only be attached at specific locations to 13 of the 23 body segment (Figure \ref{fig:teaser}). These specific attachment sites should be respected at inference as they are used to synthesize the training data (Section \ref{section_Datasets}). 

For the upper and lower limbs we opted to attach the sensors in positions as distal as possible without hampering joint motion. As location within a body segment does not influence the orientation estimate of a body, we choose this placement to maximize the signal-to-noise ratio for the measured linear acceleration. The root and torso IMU were attached at locations that provide a good articulation between sensor and body segment with few soft tissue artifacts. 
Similar to prior work \cite{Yi21,Yi22,Jiang22} we will ignore the toes, wrist and finger joints in our evaluations as in several motion training datasets these dofs are not articulated.

\subsection{Diffusion Model}

\subsubsection{Features}
Our diffusion model generates sequences of N=61 feature vectors. The feature vector in each frame represents the combination of whole-body pose and IMU information in a compact manner:
\begin{equation}
\label{eq:state}
    \bm{x}_{frame} = (\bm{R}, \bm{a}, \Delta\bm{p}, \bm{p_y}, \bm{b}).
\end{equation}
$\bm{R}\in \bm{R}^{24\times6}$ are the global orientations of the body segments, parameterized by 6-DOF representation \cite{Zhou2019}. The orientations of the 13 body segments that are potentially instrumented also represent IMU orientation estimates as there is no relative orientation between IMU and body segment after calibration. $\bm{a}\in \bm{R}^{13\times3}$ are the linear accelerations, expressed in the world frame, of the IMU locations on the potentially instrumented body segments. To complete our whole-body motion representation we have $\Delta\bm{p}$, the 2D change in root position from one frame to the next, and $\bm{p_y}$ the root vertical position. For the heel and toe of each foot, we add a binary contact feature $ \bm{b} \in \bm{R}^{2\times2} $ that will be used to regularize generated motion (Section \ref{Training procedure and losses}). The full feature vector is $ \bm{x} \in \bm{R}^{61x190}$.

\subsubsection{Diffusion Framework}

\begin{figure}
\centering
    \includegraphics[width=\linewidth]{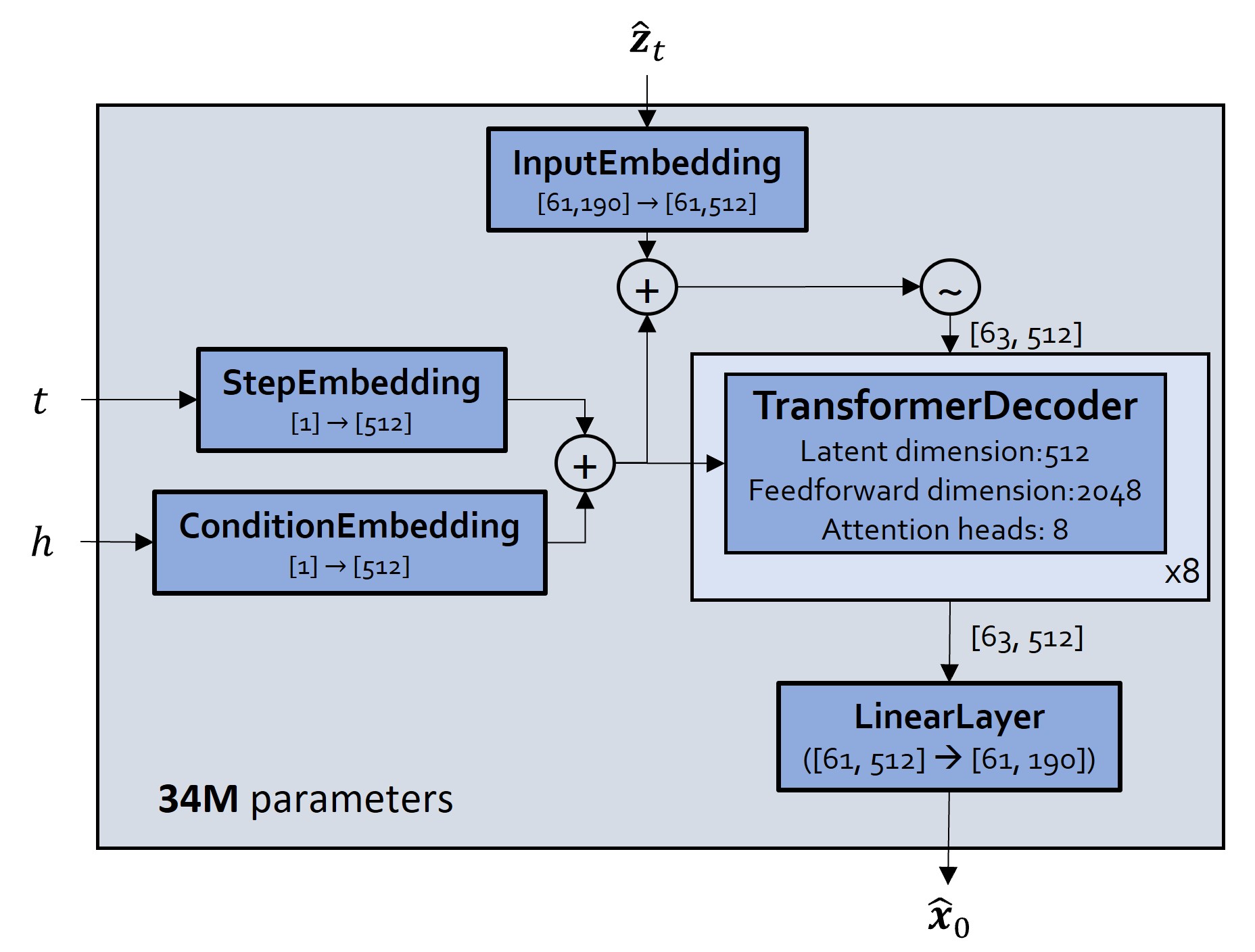}
  \caption{\textbf{DiffIP transformer decoder network.} Architecture of the denoiser $\bm{f}_{\theta}(\mathbf{\hat{z}}_t,t,h)$ that predicts the sample ($\hat{\bm{x}}_{0}$) given the noised sample $\bm{z}_t$, denoising step $t$ and the body height $h$. We use the transformer decoder architecture from \cite{Vaswani17} and use the step embedding and height embedding for cross attention as well as self attention by concatenating them to the input embedding.}
  \label{DiffIP_transformer}
\end{figure}

Training and architecture of our diffusion framework are similar to \cite{Tevet23, Tseng22}.
Diffusion is modeled as a Markov noising process with latents $\{ \bm{x}_{t} \}_{t=0:T}$, with $T=1000$. The forward noising process is defined as:
\begin{equation}
q(\bm{z}_{t}|\bm{x})\sim\mathcal{N}(\sqrt{\overline{\alpha}_t}\bm{x},(1-\overline{\alpha}_t)\bm{I})    
\end{equation}
where $\overline{\alpha}_t \in [0,1]$ are constants which follow a monotonically decreasing (cosine) schedule for increasing $t$,  such that $\bm{x}_T \sim \mathcal{N}(0,\bm{I})$.
We learn an approximation of the reverse diffusion process, i.e., the denoising process, by training a transformer neural network $\bm{f}_\theta$ (Figure \ref{DiffIP_transformer}), with parameters $\theta$, which takes a noised version $\bm{\hat{z}}_t$ of the ground truth motion $\bm{x}$, the noise step $t$ and condition $h$ and generates a denoised version $\hat{\bm{x}}_0$ that aims to match $\bm{x}$: 
\begin{equation}
\bm{f}_\theta(\bm{\hat{z}}_t,t,h) = \hat{\bm{x}}_0. 
\end{equation}
The condition $h$ is subject height. Height is associated with body segment lengths, which mathematically underlie the relation between body orientations and body accelerations and, body orientations and root motion.
\subsubsection{Training losses}
\label{Training procedure and losses}
For training $\bm{f}_\theta$ we follow DDPM \cite{Ho20} by sampling a diffusion step $t$ from a uniform distribution $U \in [0,T]$, randomly sampling a motion sequence from our training dataset, noising this motion $\bm{x}$ to $\bm{\hat{z}}_t$, predicting $\hat{\bm{x}}_{0}$ and performing gradient descent on our loss $\mathcal{L}$ 
\begin{equation}
\mathop{\mathbb{E}}_{\bm{x},t}\mathcal{L}(\hat{\bm{x}}_0,\bm{x}).
\end{equation}
Similar to \cite{Tseng22}, our loss is composed of a simple loss \cite{Ho20,Tevet23} and several auxiliary losses:
\begin{equation}
\mathcal{L} = \mathcal{L}_{\rm simple} + \mathcal{L}_{\rm vel} +  \mathcal{L}_{\rm FK} +\mathcal{L}_{\rm drift} + \mathcal{L}_{\rm slide}
\end{equation}
with 
\noindent
\begingroup
\small   
\thinmuskip=\muexpr\thinmuskip*5/8\relax
\medmuskip=\muexpr\medmuskip*5/8\relax
\begin{flalign}
& \mathcal{L}_{\rm simple} = \sum_{i=1}^N||\hat{\bm{x}}_0^{(i)} - \bm{x}^{(i)}||^2,\\
\noindent
& \mathcal{L}_{\rm vel}  = \sum_{i=1}^{N-1}||(\hat{\bm{R}}_0^{(i+1)} - \hat{\bm{R}}_0^{(i)}) - (\bm{R}^{(i+1)} - \bm{R}^{(i)})||^2,\\
\noindent
& \mathcal{L}_{\rm FK}  = \sum_{i=1}^N||{\rm FK}(\hat{\bm{R}}_0^{(i)}) - {\rm FK}(\bm{R}^{(i)})||^2,\\
\noindent
& \mathcal{L}_{\rm drift}  = \sum_{i=1}^N||\hat{\bm{p}}_0^{(i)} - \bm{p}^{(i)}||^2\;\; {\rm where}\;\;\bm{p}^{(i)}  = \sum_{j=1}^i \Delta\bm{p}^{(j)},\\
\noindent
& \mathcal{L}_{\rm slide}  = \sum_{i=1}^{N-1}||\hat{\bm{b}}_0^{(i)} \cdot [{\rm FK}_{\rm ft}(\hat{\bm{R}}_0^{(i+1)}) - {\rm FK}_{\rm ft}(\hat{\bm{R}}_0^{(i)}) +  \bm{\Delta p}_0^{(i)}]||^2.
\end{flalign}
\endgroup
$\mathcal{L}_{\rm simple}$ supervises the features, including contact labels $\hat{\bm{b}}_0^{(i)}$, directly. $\mathcal{L}_{\rm vel}$ encourages smooth motion \cite{Petrovich21}. $\mathcal{L}_{\rm FK}$ is a kinematic loss to encourage realistic joint positions \cite{Shi20}, with ${\rm FK_{\rm ft}}$ the forward kinematics functions mapping global orientations to foot positions relative to the root. Compared to prior work we explicitly add $\mathcal{L}_{\rm drift}$, which penalizes drift by accumulating the absolute root translation error across frames. $\mathcal{L}_{\rm drift}$ is required because we choose to predict $\Delta \bm{p}$, rather than $ \bm{p}$; a design choice to improve performance when generative motion models are used autoregressively \cite{Rempe21}. 
$\mathcal{L}_{\rm slide}$ encourages the model to generate motion where foot motion is consistent with foot contact prediction (minimizing foot velocity when in contact) \cite{Tseng22}.

\subsection{Training data}
\label{section_Datasets}
We used AMASS \cite{Mahmood19} for training. IMU orientations were synthesized as the global orientations of the body segments. IMU linear accelerations were derived by double differentation of selected vertex positions that were obtained by a forward kinematic pass. To better match measured and synthesized linear accelerations we averaged these at the original sampling rate using a moving average sliding window of 166ms (e.g., 11 frames at 60Hz) \cite{Jiang22}. Finally, we resampled at 20Hz. Contact labels were annotated by applying a velocity threshold (0.3m/s) for each of the four potential contact points (heels and toes) \cite{Yi21}. 

Sampling from training data was done following a probability that was assigned to each trial based on a simple energy metric. The sampling strategy encouraged the model to predict more diverse motion when deployed purely generative and improved reconstruction accuracy (Section \ref{section:ablations}). For the energy metric, we estimated center-of-mass ($COM$) position of each segment as the midpoint between joints and calculated $COM$ velocity. Then we calculated  linear kinetic energy for each segment: $0.5\cdot m_{segment} \cdot \dot{COM}^2$, with $m_{segment}$ in Appendix D.

\subsection{Inference}
\subsubsection{Autoregressive inference}\label{section:autoregressive_inference}
\begin{figure}[bht!]
\centering
    \includegraphics[width=\linewidth]{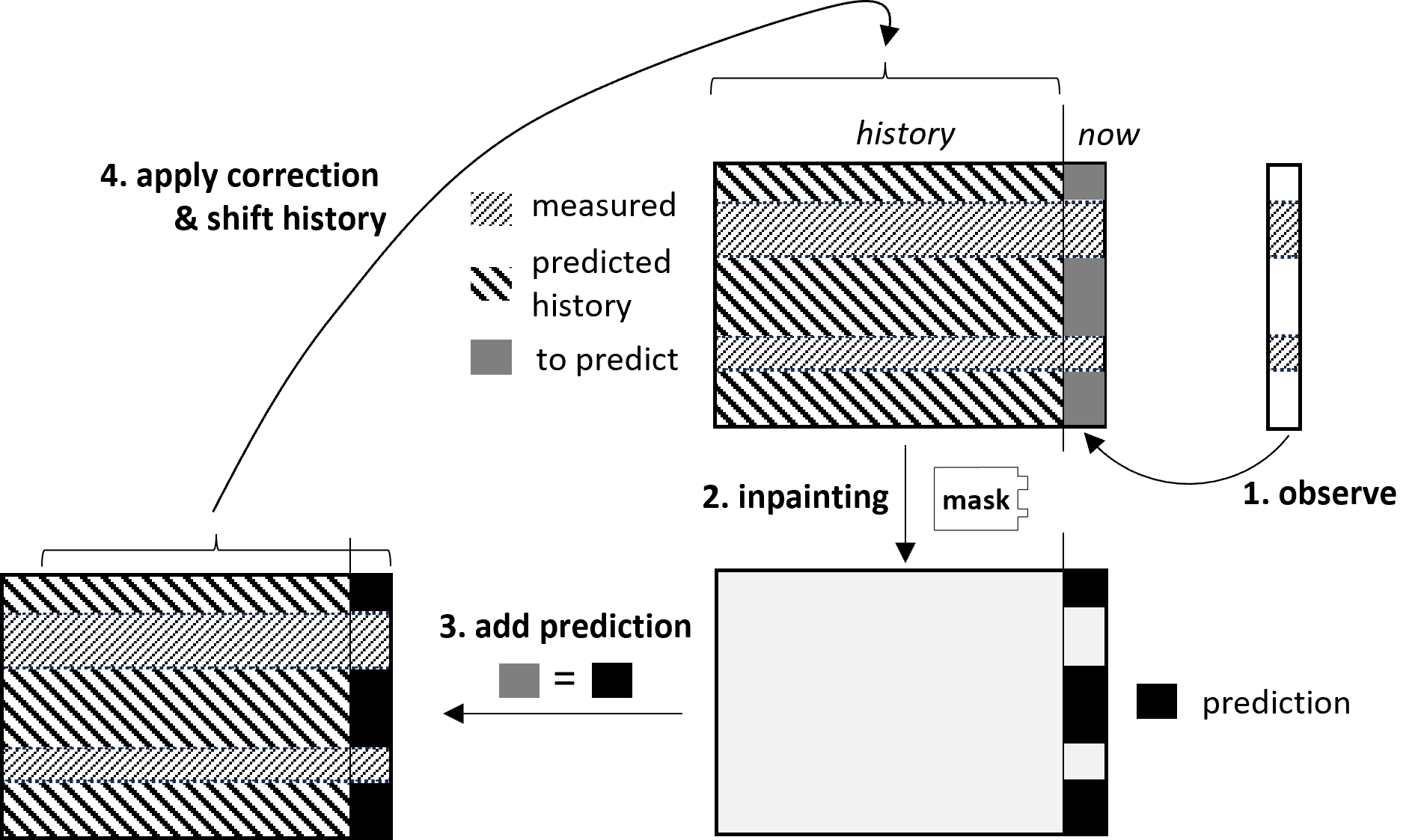}
  \caption{\textbf{Four step autoregressive inference including denoising inpainting.} Motion is reconstructed frame-by-frame in real-time following a four step process. New predictions are shifted into history and serve as input for the reconstruction at the consecutive timesteps.}
  \label{fig_InferenceProcess}
\end{figure}

Inference is run at 20Hz and consists of a four step process (Figure \ref{fig_InferenceProcess}). At each timestep a new observation is made of the features measured by the sensors. The new observation, together with the history of measured features and the history of reconstructed motion are input to the inpainting denoising process. The output from the inpainting denoising \cite{song2020denoising} process is a prediction of unobserved features at the current frame. The predicted features are concatenated with the measured features. Finally, we apply a correction of the root motion \ref{sec:root_correction} and shift the new frame into the history before a new measurement is acquired.

Our inpainting denoising algorithm \ref{inference_algorithm} takes as input $\bm{x}^{\rm input}$ which consists for frames $0$ to $N-1$ of the measured and reconstructed feature history. For the final frame, frame $N$, the observed features are set to the sensor measurements. The unobserved features of frame $N$, that we will predict, are initialized from the reconstructions at frame $N-1$. The inpainting mask ($\bm{m}$) covers all features for frames $0$ to $N-1$ and the observed features for frame $N$. $\bm{x}^{\rm input}$ are inpainted using our denoiser, which leads to $\bm{x}^{\rm output}$. From $\bm{x}^{\rm output}$ we take the unobserved features at frame $N$ and concatenate these with the measured features.

\begin{algorithm}
Given : $c, \bm{m}, \bm{x}^{\rm input}$\\
  $\bm{x}_{T} = \bm{x}^{\rm input} $ \\
  \For{t= T-1, T-2, ... 1}{
    noise $\hat{\bm{z}}_{t} \sim \mathcal{N}(\sqrt{\overline{\alpha}_t}\bm{x}_{t+1},(1-\overline{\alpha}_t)\bm{I})$ \\
    predict $\hat{\bm{x}}_0 = p_{\theta}(\hat{\bm{z}}_{t},t,c)$\\
    edit $\bm{x}_t = \bm{m} \odot \bm{x}^{\rm input} + (1-\bm{m}) \odot \hat{\bm{x}}_0$ \\
  }
  $\bm{x}^{\rm output} = \bm{x}_t$
  \caption{Inpainting denoising with DiffusionPoser}
  \label{inference_algorithm}
\end{algorithm}

\subsubsection{Root correction}\label{sec:root_correction}
We exploit the predicted contact information to correct foot sliding artifacts and reduce root drift. The motion of points on the feet that are predicted to be in contact during the transition from frame $N-1$ to $N$ is corrected by changing the root motion. We do this by first calculating the mean horizontal position change between frames $N-1$ and $N$ across the points that are predicted to be in contact. We then subtract this mean position change from the predicted horizontal root position change between frames $N-1$ and $N$. Note that this heuristic does not guarantee predicted contact points to be static, as it only makes the mean of the contact points to be static.

\section{Evaluations} 

We performed several experiments and evaluations to quantify reconstruction performance for different IMU configurations, to compare DiffusionPoser to Transpose \cite{Yi21}, PIP \cite{Yi22} and TIP \cite{Jiang22} and to demonstrate multimodality. We show that DiffusionPoser has a comparable reconstruction accuracy to other methods, while offering users the flexibility to select sensor setting on-the-fly. Next, we performed experiments to show robustness of DiffusionPoser to sensor signal corruption and to examine the effect of different denoising schemes. Finally we did several ablations.

Following \cite{Yi21,Yi22,Jiang22} we performed evaluations using the TotalCapture dataset with real IMU data for sensors on the pelvis, head, wrists and shanks (`TotalCaptureReal')\cite{Trumble17, Huang18}. 
Next we also used `TotalCaptureSynth', a version of the TotalCapture dataset where we synthesized all IMU signals. `TotalCaptureSynth' does not allow to evaluate sim-to-real error but is useful to understand the effect of different configurations on reconstruction accuracy. We also used DIP-IMU for additional baseline comparison \cite{Huang18}. Evaluations are done using 30 denoising steps (Section \ref{reduce_denoising_steps}).

We report the following evaluation metrics:\newline
\textbf{Local Angular Error, LA} [$^{\circ}$]: Rotation difference between ground-truth and reconstructed local joint angles. We use specific LA metrics to quantify reconstruction accuracy of the back (neck, shoulder and spine joints) and legs (hip, knee and ankle joints).\newline
\textbf{Global Angular Error, GA} [$^{\circ}$]: Rotation difference between ground-truth and reconstructed local global segment orientations.\newline
\textbf{Joint Position Error, JPE} [$cm$]: Difference between corresponding joint positions, expressed in the root frame, of reconstructed and ground-truth motions.\newline
\textbf{Jitter} [$-$]: Ratio of the global jitter averaged across joints of the reconstructed over the jitter of the ground truth motion. Global jitter of a joint was calculated as the third derivative of absolute position using finite differences. \newline
\textbf{Root Translation Error, RE} [$m$]: Distance between root position of ground-truth and reconstructed motion at 2s, 5s and 10s into the motion.

The metrics are based on Transpose, TIP and PIP. Because there is no exact correspondence between metrics across papers, values do not exactly correspond to the original papers. This is further explained in Appendix B.

\begin{figure}[bht!]
  \includegraphics[width=\linewidth]{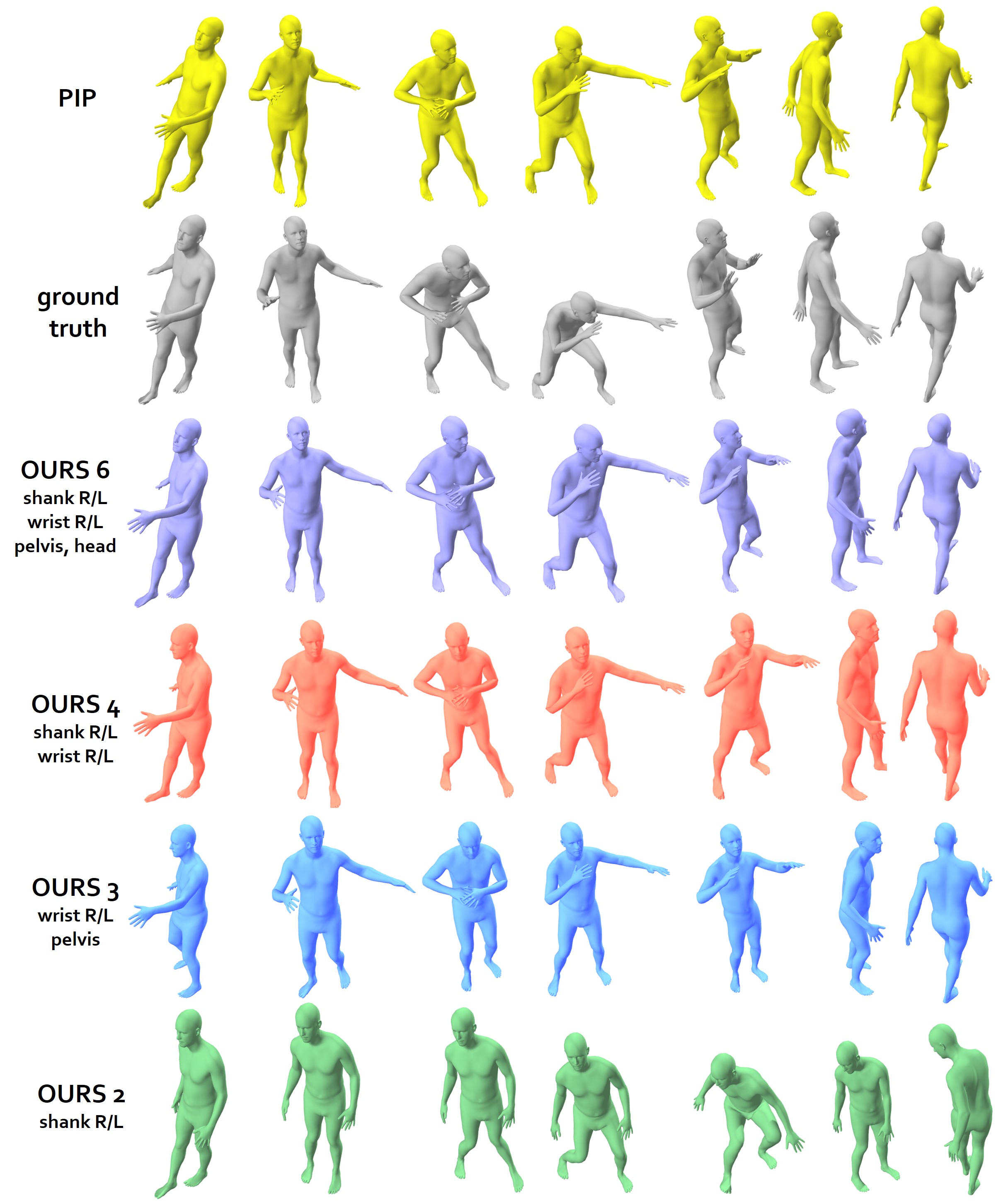}
  \caption{\textbf{Motion reconstructions with PIP and DiffusionPoser (Ours) for different IMU configurations of a TotalCaptureReal sequence.} Yellow: PIP with pelvis, head, wrists and shanks. Grey: ground truth. Purple: Ours with pelvis, head, wrists and shanks. Orange: Ours with wrists, shanks. Blue: Ours with pelvis and wrists. Green: Ours with shanks.}
  \label{fig:different_config}
\end{figure}

\subsection{Optmizing IMU configurations across tasks}
We show the capabality of DiffusionPoser to optimize sensor configurations for specific activities by evaluating on different metrics to quantify accuracy for: whole-body (GA), leg kinematics (legsLA), back kinematics (backLA) and global translation (RE10).
We optimized sensor configurations for two, three and four IMU sensors by evaluating the performance for a range of selected configurations on TotalCaptureSynth. This was the only experiment where we used TotalCaptureSynth because we did not have real IMU data for all the potentially instrumented segments that we wanted to test. 
A full table with results can be found in Appendix C, Table 4. Table \ref{sensor_config_2_3IMUs} shows the optimal configuration when using four, three and two IMU sensors for the different evaluation metrics. Note how instrumenting thighs is best to reconstruct leg kinematics, whereas shanks are preferred to improve root translation error. 
\begin{table}[hbt!]
 \caption{\textbf{Optimal IMU configurations} for four, three and two sensors for different tasks.}
 \label{sensor_config_2_3IMUs}
 {\fontsize{8}{8}\selectfont
\begin{tabular}{ p{0.1cm}p{1.55cm}p{1.55cm}p{1.55cm}p{1.55cm}}
 \toprule
 \textbf{$\#$} & GA [$^{\circ}$]&legsLA [$^{\circ}$] & backLA [$^{\circ}$] &  RE10 [m]\\
\midrule
\midrule
 $4$  & $plvs,hd, $ $wr_r,wr_l $&  $plvs, hd$ $ th_r, th_l $ & $plvs, hd $ $ arm_r, arm_l$   & $th_r, th_l$ $ sh_r, sh_l  $  \\
 \midrule
 $3$  & $plvs, $ $ wr_r, wr_l $  & $plvs, $ $ th_r, th_l $  & $plvs, $ $ arm_r, arm_l$   & $plvs, $ $ sh_r, sh_l  $ \\
 \midrule
 $2$ & $arm_r, arm_l$ & $plvs, sh_r  $ & $arm_r, arm_l $ & $sh_r, sh_l $
\\
\bottomrule
\end{tabular}}
\end{table}

\subsection{Sim-to-real error and reconstruction quality}
We evaluated real vs synthetic reconstruction quality. We performed exhaustive evaluation for all possible IMU configurations with six or fewer sensors with the potentially instrumented segments limited to wrists, shanks, pelvis and head.
Qualitative results of reconstructions from real data and a comparison to PIP \cite{Yi22} are shown in Figure \ref{fig:different_config}. 
Table \ref{sim_to_real} reports sim-to-real comparisons for selected configurations. Quantitative results for all configurations we tested with real IMU data are in Appendix C, Table 5.
\begin{table}[bht!]
\centering
 \caption{\textbf{Sim-to-real error} for selected IMU configurations.}

 \label{sim_to_real}
 {\fontsize{7}{9}\selectfont
 \begin{tabular}{
    p{1.6cm}
    p{0.7cm}
    p{0.8cm}
    p{0.9cm}
    p{0.9cm}
    p{0.9cm}
  }
  \toprule
  {\textbf{config}} & \begin{tabular}{l}{IMU}\end{tabular} & \begin{tabular}{l}{GA [$^{\circ}$]}\end{tabular}  & \begin{tabular}{l}{Jitter[-]}\end{tabular}  &\begin{tabular}{l}{RE2[m]}\end{tabular} & \begin{tabular}{l}{RE10[m]}\end{tabular}\\
   \midrule
   \midrule
   \begin{tabular}{l} $plvs, hd, wr_r, $ \\$ wr_l, sh_r , sh_l$\end{tabular} &
   \begin{tabular}{l} real \\synth\end{tabular} & \begin{tabular}{l} 14.4 \\7.0\end{tabular} &\begin{tabular}{l} 2.8 \\2.8\end{tabular}& \begin{tabular}{l} 0.25 \\0.09\end{tabular} & \begin{tabular}{l} 0.25 \\0.17\end{tabular}\\

  \midrule
     \begin{tabular}{l} $plvs, hd, $ \\$ sh_r , sh_l$\end{tabular} &\begin{tabular}{l} real \\synth\end{tabular} & \begin{tabular}{l} 24.9 \\22.3\end{tabular} &\begin{tabular}{l} 2.7 \\2.5\end{tabular}& \begin{tabular}{l} 0.26 \\0.17\end{tabular} & \begin{tabular}{l} 0.26 \\0.17\end{tabular}\\
  \midrule
     \begin{tabular}{l} $plvs, sh_r , sh_l$\end{tabular} &\begin{tabular}{l} real \\synth\end{tabular} & \begin{tabular}{l} 36.4 \\28.1\end{tabular} &\begin{tabular}{l} 2.8 \\2.5\end{tabular}& \begin{tabular}{l} 0.30 \\0.21\end{tabular} & \begin{tabular}{l} 0.30 \\0.21\end{tabular}\\
  \midrule
     \begin{tabular}{l} $sh_r , sh_l$\end{tabular} &\begin{tabular}{l} real \\synth\end{tabular} & \begin{tabular}{l} 39.2 \\29.9\end{tabular}  &\begin{tabular}{l} 4.3 \\3.1\end{tabular}&  \begin{tabular}{l} 0.47 \\0.26\end{tabular} &  \begin{tabular}{l} 0.47 \\0.26\end{tabular}\\
   \bottomrule
 \end{tabular}}
\end{table}

\begin{table*}
\caption{ \textbf{Comparison of DiffusionPoser to different baselines for TotalCaptureReal}. Numbers are averages over all trials, bracketed numbers are the metrics from the trial with the highest error.}
\label{evaluate_vs_baselines}
 {\fontsize{8}{8}\selectfont

\begin{tabular}{  p{1.7cm}p{1.7cm} p{1.7cm}p{1.7cm}p{1.7cm}p{1.7cm}p{1.7cm}p{1.8cm}}
\toprule
\textbf{system}& LA [$^{\circ}$] & GA [$^{\circ}$] & JPE [cm] & Jitter [-] & RE2s [m] & RE5s [m] & RE10s [m]\\  
\midrule
\midrule
  Transpose & $13.9(24.3)$ & $16.1(25.2)$ & $6.4(12.7)$ & $5.0(9.7)$ & $0.19(0.98)$ & $0.26(1.38)$ & $0.29(1.4)$ \\
\midrule
PIP & $11.9(20.1)$ & $14.4(22.7)$ & $5.3(10.7)$ & $1.1(1.5)$ & $0.12(0.45)$ & $0.17(0.60)$ & $0.27(1.07)$ \\  
\midrule
  TIP & $12.0(20.0)$ & $14.3(21.4)$ & $6.2(12.1)$ & $5.3(13.0)$ & $0.17(0.46)$ & $0.33(0.53)$ & $0.32(0.98)$ \\
\midrule
DiffusionPoser & $13.0(21.5)$ & $14.4(23.4)$ & $6.1(12.3)$ & $2.8(4.5)$ & $0.14(0.41)$ & $0.20(0.6)$ & $0.25(0.75)$\\
\bottomrule
\end{tabular}}
\end{table*}
Despite the many sources of sim-to-real-error our model performs well on real data (see also Section \ref{subsection:compare_baselines}). The following are sources that contribute to sim-to-real error: (1) imperfect sensor-to-bone orientation calibration, (2) relative motion between sensor and bone, (3) imperfect placement of the IMU with respect to vertex positions used during data synthesis, (4) IMU orientation estimation error by the proprietary algorithm and (5) differences synthetic and real data noise properties. We noted that (1) and (2) are substantial in `TotalCaptureReal'.

\subsection{Comparison to baselines for six IMUs}\label{subsection:compare_baselines}

We compare DiffusionPoser to Transpose, PIP and TIP on `TotalCaptureReal' for the six IMU sensor configuration: pelvis, head, wrists, shanks (Table \ref{evaluate_vs_baselines}).
We reran and evaluated Transpose, PIP and TIP to ensure we were making a fair comparison starting from the exact same input data and reporting the same evaluation metrics.

Evaluation metrics for all four systems are close. DiffusionPoser is within 1.1 degrees and 1cm of the system with the best scores for angular error and joint position error. Optical motion capture will result in variations that are similar or even larger than 1 degree for the exact same underlying motion performed by the same subject due to marker noise, differences in marker placement and parameter settings of the motion capture processing pipeline \cite{CAPPOZZO199690, WILKEN2012301}.
The jitter metric is significanlty lower in PIP than in DiffusionPoser. Several trials reconstructed by PIP have less jitter than the ground truth motion, indicating that PIP could be oversmoothing in such cases. From our videos it is clear that our reconstructed motion smoothness is reasonable. 

We performed an additional validation with real IMU data using the DIP-IMU dataset \cite{Huang18} with comparison to Transpose, PIP and TIP. Conclusions are the same with evaluation metrics being even closer as for TotalCaptureReal. Results are in Appendix Table 3.

\subsection{Multimodality: Adding shoe insoles}
Because we use contact information while reconstructing motion we can exploit ground truth information when pressure insoles are worn. We performed a quantitative analysis on `TotalCaptureReal' by using ground truth contact labels as part of the measured signals.
Using ground truth contact labels improved accuracy (Table \ref{insoles}). When wearing IMUs on the shank the improvement from wearing insoles is less because the shank IMUs provide good information to estimate the contact label.  
\begin{table}
 \caption{\textbf{Adding insoles for ground truth contact labels improves reconstruction accuracy.}}
 \label{insoles}
  {\fontsize{8}{8}\selectfont
\begin{tabular}{ p{1.2cm}p{0.8cm}p{0.8cm}p{0.9cm}p{0.9cm}p{1.1cm}}
 \toprule
 \textbf{config} & \textbf{insole} & GA[$^{\circ}$] & Jitter[-] & RE2[m] & RE10[m]\\
\midrule
\midrule
 $plvs, hd,$ $sh_r, sh_l$ & No  Yes& $29.4 $ $ \bm{29.1}$   & $2.7 $  \hspace{0.5cm}  $  \bm{2.6}$ & $0.14 $  $ \bm{0.12} $  & $0.33 $  $\bm{0.26}$\\
\midrule

 $plvs, hd,$ $wr_r, wr_l$  & No  Yes& $18.0$ $\bm{16.9}$   &   $3.2 $ \hspace{0.5cm} $ \bm{3.0} $  &$0.34 $ $ \bm{0.3} $& $0.96$ $ \bm{0.67} $     \\
\bottomrule
 \end{tabular}}
\end{table}

\subsection{Dealing with signal corruption and loss}\label{subsection:signal_loss}
DiffusionPoser can deal with corrupted or lost signal by relying on its generative nature. Signal loss and corruption occurs regularly in practical settings (e.g. packet loss, out-of-range). We provide examples on our project website where we drop the signal from all sensors for a couple of seconds. DiffusionPoser continues to generate realistic motion in real-time. Once the signal is back, a natural transisition to the actual motion is generated. Regressive models, such as PIP \cite{Yi22} are not capable of such online infilling.

\subsection{Reducing denoising steps and optimal spread}
\label{reduce_denoising_steps}
For DiffusionPoser to run online we sped up the denoising process by reducing the number of denoising steps following DDIM \cite{song2020denoising,Du23}. For our large base model, a NVIDIA A4000 GPU can achieve real-time at 30 denoising steps, but for weaker GPUs it is required to reduce to 10 or even 5 denoising steps. Here we analyzed how accuracy degrades with reducing the number of denoising steps (Table \ref{denoising_steps}) and how these steps are best spread (Table \ref{denoising_config}).

Accuracy improved with increasing number of denoising steps but saturated around 30 steps. Scaling up to 100 denoising steps only gave a marginal improvement. When reducing the number of denoising steps to 5 we sacrificed the jitter metric mostly.

Concerning different denoising schemes (Table \ref{denoising_config}), for 10A we only used the last 10 denoising steps. For 10B we used the last 20 denoising steps with a spread of 2. For 10C we started at step 100 and decayed exponentially to zero. 10D has the following irregular spread: 1000/850/700/550/400/250/100/10/2/0.
We observed a trade-off where starting from high noise levels resulted in slightly worse angular errors and more jitter but improved root estimation error. This could be explained as we initialized the unknown features of the last frame as a copy of the prior frame. Since orientations are smooth, a few denoising steps are enough for the model to make a good prediction of the next frame. The change in root position ($\Delta p$) is not smooth and adding more noise allows the model to change the initialization of the last frame more. For all other evaluations in this paper we used a spacing similar to 10D as the increase in angular error and jitter was small relative to the gain in root estimation accuracy.

\begin{table}
 \caption{\textbf{Evaluation of decreasing numbers of denoising steps for the 6 IMU configuration.}}
 \label{denoising_steps}
  {\fontsize{8}{8}\selectfont
\begin{tabular}{ p{1cm}p{1.20cm}p{1.15cm}p{1.4cm}p{1.4cm}}
 \toprule
 $\mathbf{\#}$\textbf{steps} & GA[$^{\circ}$] & Jitter[-] & RE2[m] & RE10[m]\\
\midrule
\midrule
 5 & $16.0$ & $4.4$  &  $0.17 $ &  $ 0.32$  \\
 \midrule
 10 & $ 15.8$ & $3.3$ &$ 0.15 $  &$ 0.25$  \\
 \midrule
 30 & $14.4$  & $2.8$  & $\bm{0.14} $&   $0.25$    \\
 \midrule
 100 & $ \bm{14.3}$  & $\bm{2.8}$  & $0.14$   & $\bm{0.24}$ \\
 \bottomrule
 \end{tabular}}
\end{table}

\begin{table}
 \caption{\textbf{Evaluation of differently spreading denoising steps.}}
 \label{denoising_config}
  {\fontsize{8}{8}\selectfont
\begin{tabular}{ p{1cm}p{1.20cm}p{1.15cm}p{1.4cm}p{1.4cm}}
\toprule
\textbf{spread} & GA[$^{\circ}$] & Jitter[-] & RE2[m] & RE10[m]\\
 \midrule
 \midrule
 10A & $\bm{15.3}$  & $\bm{2.0}$  &  $0.42$  &   $0.97$  \\
 \midrule
 10B & $15.4$  & $2.1$  & $0.42$   & $0.89 $ \\
  \midrule
 10C & $15.4$  & $2.4 $ &$ 0.45 $& $  0.67  $  \\
  \midrule
 10D & $15.8$  & $3.3 $ & $\bm{0.15} $  & $\bm{0.25} $\\
\bottomrule
 \end{tabular}}
\end{table}

\subsection{Additional experiments and ablations}\label{section:ablations}
We performed several additional experiments and ablations (Table \ref{table_AblationAndModelSize}). We found that reducing the number of transformer layers is a more effective way to reduce model size than reducing the feature dimension. Interestingly, and we do not know why, root translation error was slightly better in smaller models. 
We found that not using the energy metric for sampling during training increased the GA by 1$^{\circ}$. Next, we found that ablating height as a condition did not change much. This could in part be explained by the subjects in TotalCapture having an average stature.
Finally, it is clear that using the contact label predictions to perform root motion correction improves root translation error. Root correction slightly increased jitter. 

We performed a comparison of DiffusionPoser with a regressive model that takes in an input mask to encode sensor configuration. This was only implemented for the OpenSim version of DiffusionPoser and we show that DiffusionPoser is more accurate and robust than the regressive model (Appendix A, Table 2).

\begin{table}
\caption{ \textbf{Ablations}. Transformer model size is reported between brackets: layers/feature dimension/feedforward dimension}
\label{table_AblationAndModelSize}
 {\fontsize{8}{8}\selectfont
\begin{tabular}{ p{2.2cm}p{0.8cm}p{1.1cm}p{1.0cm}p{1.1cm}}
 \toprule
  \textbf{Ablation} & GA[$^{\circ}$] & Jitter[-]& RE2[m] & RE10[m]\\
  \midrule
  \midrule
  Ours\\ (8/512/2048)  &$14.4$  & $2.8$ & 0.14 & $0.25$ \\
  \midrule
   model size \\(4/512/2048)& $ 14.9$ & $ 2.7 $& 0.11 & $ 0.22 $ \\
  \midrule
  model size \\(8/256/1024) & $ 15.3$ & $ 2.7 $& 0.11 &  $ 0.25 $ \\
  \midrule
  Ours \\
  w/o energy metric  &  $ 15.4  $ & $ 2.6$ & 0.14&  $  0.28$\\
  \midrule
  Ours \\
 w/o height    & $ 14.5$ & $ 2.8$ & 0.14& $ 0.27 $\\
  \midrule
  Ours \\
    w/o root correction  & $ 14.4 $ & $ 2.15 $& 0.32 &  $ 0.54$\\
  \bottomrule
\end{tabular}}
\end{table}

\subsection{Live demonstration}
We tested our system live with different configurations in an indoor and outdoor setting (tennis court). Using our system requires a quick sensor-to-world calibration. Next, the sensors are attached at the correct body locations and a sensor-to-bone calibration is done by standing in T-pose.

We use XSens IMU sensors and stream processed (MTw Awinda) orientation and linear acceleration at 60Hz. We apply a moving average filter with current frame and five past and five future frames on the acceleration and downsample both signals to 20Hz to match our synthetic training data.

We perform indoor experiments with a mixture of motions including walking, jogging, jumping, lunges and kicking and some specific gait deviations that are observed in patient populations. Next we also perform a live demonstration of our system with a tennis player outdoors.

The latency between the real and visualized motion results from (1) XSens to system communication, (2) moving average filter (83ms) and (3) our reconstruction algorithm (45ms). We noticed that the latency of the XSens system increased when the distance to the receiver station increased. The latency of our reconstruction algorithm is similar to TIP \cite{Jiang22}, and significantly more than PIP (16ms) \cite{Yi22}. 

\section{Conclusion}
DiffusionPoser allows immediate use of arbirtary sensor configurations and thus optimizing these configurations for specific activities. This generative model provides robustness in motion prediction against sparsity, noisy and corrupted sensor measurements. Our evaluations show state-of-the-art performance when using six sensors and little degradation in performance when using fewer sensors. 

The latency of DiffusionPoser is a limitation for applications that require short latencies such as visual illusion ($<$50ms according to \cite{Stauffert20}) or control of an assistive device (e.g., 40-60ms to assist balance with an exoskeleton according to \cite{Beck23}). 
Other future work is to extend our system to estimate joint torques and muscle forces. This could lead to a breakthrough system for biomechanists and users interested in health and performance.

{\small
\bibliographystyle{ieee_fullname}
\bibliography{egbib}
}
\clearpage
\end{document}